\documentclass[conference]{IEEEtran}
\IEEEoverridecommandlockouts

\usepackage{amsmath,amssymb,amsfonts, mathtools}
\newcommand{\colvec}[2][.9]{%
  \scalebox{#1}{%
    \renewcommand{\arraystretch}{1}%
    $\begin{bmatrix}#2\end{bmatrix}$%
  }
}
\usepackage{algorithm}
\usepackage{algorithmic}
\usepackage{graphicx}
\usepackage{textcomp}
\usepackage{colortbl}
\usepackage[table,xcdraw]{xcolor} 
\usepackage{array}
\usepackage{multirow}
\usepackage{makecell}
\usepackage{hyperref}
\usepackage[normalem]{ulem}
\usepackage{siunitx}
\usepackage{mathrsfs}
 \usepackage{booktabs}
 \usepackage{enumitem}

\usepackage{amsthm}
\theoremstyle{remark}
\newtheorem*{remark}{Remark}

\usepackage{tikz}
\usepackage{tikz-3dplot}

\usepackage{svg}

\ifCLASSOPTIONcompsoc
\usepackage[caption=false,font=normalsize,labelfont=sf,textfont=sf]{subfig}
\else
  \usepackage[caption=false,font=footnotesize]{subfig}
\fi

\definecolor{lightgray}{gray}{0.9}
\definecolor{darkgreen}{RGB}{0,135,60}
\definecolor{darkorange}{RGB}{210,110,0}

\usepackage[
placement=top,
angle=0,
color=red,
scale=1.1,
hshift=0,
vshift=-15
]{background}
\backgroundsetup{contents=\bf{----------------------------------------------------------------- PREPRINT -----------------------------------------------------------------}}
\begin{document}


\title{
Force Polytope–Based Cant--Angle Selection\\ for Tilting Hexarotor UAVs
\thanks{*This work is partially financed by the European Union - Next Generation EU, Mission 4 Component 2 - CUP C53D23000520006 STARLIT - SafeTy Aware Reinforcement Learning for robotIc inspecTion.}
}

\author{\IEEEauthorblockN{Alberto Piccina$^1$, Massimiliano Bertoni$^1$, Angelo Cenedese$^{2,3}$, Giulia Michieletto$^{1,2}$}
\IEEEauthorblockA{\textit{$^1$Department of Management and Engineering}, \textit{University of Padova},
Vicenza, Italy  \\
\textit{$^2$Department of Information Engineering}, \textit{University of Padova},
Padova, Italy\\
\textit{$^3$Department of Industrial Engineering}, \textit{University of Padova},
Padova, Italy\\
corresponding contact: \texttt{alberto.piccina@phd.unipd.it}}
}

\maketitle

\begin{abstract}
From a maneuverability perspective, the main advantage of tilting multirotor UAVs lies in the dynamic variability of the feasible executable wrench, which represents a key asset for physical interaction tasks.
Accordingly, cant-angle selection should be optimized to ensure high performance while avoiding abrupt variations and preserving real-world feasibility.
In this context, this work proposes a lightweight control framework for star-shaped interdependent cant-tilting hexarotor UAVs performing interaction tasks. The method uses an offline-computed look-up table of zero-moment force polytopes to identify feasible cant angles for a desired control force and select the optimal one by balancing efficiency and smoothness.
The framework is integrated with a geometric full-pose controller and validated through Monte Carlo simulations in MATLAB/Simulink and compared against a baseline strategy. The results show a significant reduction in computation time, together with improved pose-tracking performance and competitive actuation efficiency. 
A final physics-based simulation of a complete wall inspection task in Simscape further confirms the feasibility of the proposed strategy in interacting scenarios.

\end{abstract}



\section{Introduction}\label{sec:introduction}
Recent advances in aerial robotics have enabled multirotor platforms to perform physical interaction tasks such as contact-based inspection, aerial grasping, and object manipulation~\cite{ruggiero2018aerial,ollero2021past,piccina2025taxonomy} and references therein. 
Interactive tasks generally require multirotor platforms to generate desired interaction forces and moments in the three-dimensional space, while simultaneously performing trajectory tracking. 

The ability of a multirotor to generate interaction forces is fundamentally constrained by its actuator configuration. Traditional platforms with fixed thrust directions are intrinsically underactuated, which limits their capability to produce arbitrary wrenches~\cite{idrissi2022review}. In contrast, thrust-vectoring architectures enable improved force–moment decoupling.
Among these, tilted multirotor UAVs provide enhanced maneuverability and improved force–moment decoupling, making them well-suited for interaction tasks~\cite{ryll20176d}. These platforms are typically equipped with propellers whose spinning axes are rotated around the corresponding arms by a constant (cant) angle, which is selected to guarantee full actuation while optimizing energy efficiency, particularly during position and trajectory tracking.
Although less common, tilting multirotor UAVs extend this concept by allowing propeller spinning axes direction to be reconfigured during flight through one or more additional degrees of freedom~\cite{ryll2012modeling,ryll2021fast,kamel2018voliro}. By adapting the cant angles, these systems dynamically reshape their feasible wrench space, enabling both energy-efficient free flight and precise in-contact operations.

The increased maneuverability enabled by time-varying cant tilt angles comes at the cost of greater mechanical and control complexity. Specifically, additional degrees of freedom must be explicitly controlled, and their online selection is generally strongly dependent on the application requirements. In physical interaction scenarios, it is therefore natural to adapt the tilt angles with the dual objective of accurately achieving the desired interaction force while limiting the associated computational burden. From a control perspective, this problem is typically addressed within the control allocation framework, which aims to determine the most suitable actuator commands (in terms of both cant tilt angles and spinning rates) corresponding to a given desired wrench.

Control allocation for tilting multirotor platforms has been addressed through different strategies~\cite{johansen2013control}. A widely used approach relies on the pseudo-inverse of the allocation matrix, which assumes a linear mapping from the desired wrench to actuator commands. For example, in~\cite{kamel2018voliro}, control allocation for the Voliro platform is performed via the Moore–Penrose pseudo-inverse, minimizing control effort by leveraging the instantaneous actuation geometry.  However, this method does not explicitly consider actuator constraints or nonlinearities, which can limit performance during complex maneuvers or interactions.
More sophisticated methods jointly allocate propeller spinning rates and tilt angles by formulating constrained optimization problems. For instance, the MOMAV platform~\cite{ruggia2025momav} employs a Sequential Quadratic Programming (SQP) approach to optimize control inputs under constraints. While this improves actuator utilization and feasibility, such optimization-based methods are computationally intensive and may not be suitable for real-time applications with fast dynamics. Alternatively, some platforms, like Fast-Hex~\cite{ryll2021fast}, simplify the problem by fixing the cant angle and manually tuning it offline according to task requirements. This reduces computational load but sacrifices adaptability and optimality in dynamic scenarios, as the control allocation assumes a fixed rotor geometry regardless of changing conditions.

In this work, we focus on the control of a specific class of star-shaped interdependent cant-tilting hexarotors. These platforms are equipped with six propellers equally spaced along a circumference centered at the vehicle’s center of mass (CoM). The propellers’ spinning axes can tilt during flight; however, consecutive propellers are constrained to tilt by the same magnitude but in opposite directions.
This design preserves overall geometric symmetry while allowing the UAV to reshape the space of feasible executable wrench directions online through a single collective tilt parameter. As a result, it provides a favorable trade-off between improved maneuverability and mechanical and algorithmic complexity.
The control framework proposed for these UAVs relies on the geometric characterization of the executable force as a function of the assignable cant angle and aims to reduce the computational burden associated with tilt-angle allocation during interaction tasks. Specifically, we propose an approach that leverages:
\begin{itemize}
\item an offline-computed look-up table (LUT) that maps the tilt angle to the corresponding three-dimensional polytope of executable force, according to the geometric interpretation and analysis outlined in~\cite{perin2024star};
\item an online candidate cant--angle identification step based on a safety--margin containment condition for a desired control force;
\item an online optimization of the selected cant angle among feasible candidates.
\end{itemize}

The key idea is to shift most of the geometric feasibility analysis offline, so as to enable lightweight online cant-angle selection without resorting to computationally demanding joint nonlinear allocation.

The remainder of the paper is organized as follows.
Section~\ref{sec:model_and_polytope} summarizes the system model and the formulation of the force polytope.
Section~\ref{sec:my_method} details the proposed control strategy for tilt-angle selection.
Section~\ref{sec:validation} presents the performance assessment of the method via MATLAB/Simulink environment.
Finally, Section~\ref{sec:conclusion} draws the conclusions.


\section{{Modeling Preliminaries}}\label{sec:model_and_polytope}
This section lays the foundations for the proposed cant-angle selection strategy, providing a comprehensive description of the star-shaped interdependent cant-tilting hexarotors and of the corresponding force polytope used to characterize their actuation capabilities, according to~\cite{perin2024star}.

\subsection{Star-shaped Interdependent Cant-Tilting  Hexarotors}
\begin{figure}[!t]
    \centering
    \includegraphics[width=0.75\linewidth]{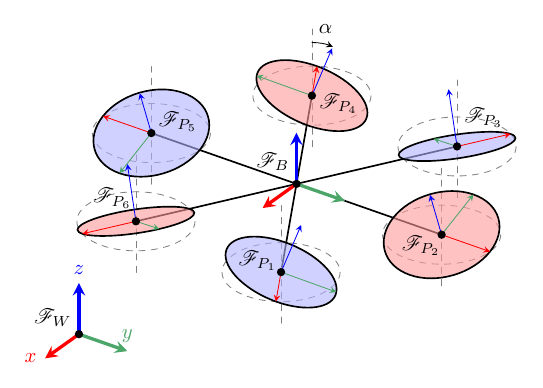}
    \caption{Schematic representation of a star-shaped interdependent cant-tilting hexarotor. 
    Each propeller spinning axes can be online reoriented by selecting the cant angle $\alpha$, thus enabling reconfiguration of the achievable wrench space.
    }
    \label{fig:FA6}
\end{figure}

To model the dynamics of a star-shaped interdependent cant-tilting hexarotor (Figure~\ref{fig:FA6}), we introduce the global inertial frame $\mathscr{F}_W = \{ O_W,\; (\mathbf{x}_W,\, \mathbf{y}_W,\, \mathbf{z}_W) \}$ (\textit{world frame}) whose axes directions are identified by the unit vectors $\mathbf{e}_1$, $\mathbf{e}_2$ and $\mathbf{e}_3$ of the canonical basis of $\mathbb{R}^3$ for the sake of simplicity, and the frame $\mathscr{F}_B = \{ O_B,\; (\mathbf{x}_B,\, \mathbf{y}_B,\, \mathbf{z}_B) \}$ (\textit{body frame}), attached to the vehicle's CoM.  In addition, each $i$-th propeller, $i \in \{ 1\dots6\}$ is associated with a local frame $\mathscr{F}_{P_i} = \{ O_{P_i},\; (\mathbf{x}_{P_i},\, \mathbf{y}_{P_i},\, \mathbf{z}_{P_i}) \}$ (\textit{propeller frame}), whose origin $O_{P_i}$ coincides with the propeller spinning center, and whose axis $\mathbf{z}_{P_i}$ identifies the spinning axis.

Each propeller spinning axis can tilt over time. In detail, each local frame $\mathscr{F}_{P_i}$ can rotate around its $\mathbf{x}_{P_i}$ axis through a time-varying angle defined as $\alpha_{P_i}(t) = (-1)^i\, \alpha(t) \quad \text{with} \quad \alpha(t) \in \Gamma_\alpha=\left[ -\frac{\pi}{3}, \frac{\pi}{3}\right)$, producing an alternating tilt pattern between adjacent rotors. The time-varying parameter $\alpha(t)$ enables dynamic modification of the achievable wrench space of the star-shaped interdependent cant-tilting hexarotor (hereafter, $\alpha$-TingH).

While rotating at a spinning rate $\omega_{P_i}(t) \in [0, \bar{\omega}]$ with $\bar{\omega} \in \mathbb{R}_{\geq 0}$, each propeller generates a thrust force $\mathbf{f}_{P_i}(t) \in \mathbb{R}^3$ and a drag moment $\boldsymbol{\tau}_{P_i}^d(t) \in \mathbb{R}^3$ both aligned with its $\mathbf{z}_{P_i}$ axis
\begin{align}
   \mathbf{f}_{P_i}(t) = c_f \omega_{P_i}(t)^2 \mathbf{z}_{P_i} \quad \text{and} \quad 
 \boldsymbol{\tau}_{P_i}^d(t) = \kappa_i c_{\tau} {\omega}_i(t)^2 \mathbf{z}_{P_i},
\end{align}
where $c_f , c_\tau \in \mathbb{R}_{\geq 0}$ are constant coefficients depending on the propellers geometrical and aerodynamical characteristics, and $\kappa_i = (-1)^i$ distinguishes the clockwise and counterclockwise spinning directions\footnote{Note that we account for platforms actuated by a
set of rotors with uniform actuation and equal aerodynamic
characteristics, as well as a balanced choice of CW/CCW
spinning directions.}. Then, the control force $ \mathbf{f}_c^B(t) \in \mathbb{R}^3$ and the control moment $\boldsymbol{\tau}_c^B(t) \in \mathbb{R}^3$  acting on the platform body frame depend on the cant angle $\alpha(t)$ as
\begingroup
\setlength{\abovedisplayskip}{3pt}
\setlength{\belowdisplayskip}{3pt}
\setlength{\abovedisplayshortskip}{4pt}
\setlength{\belowdisplayshortskip}{4pt}
\begin{align}
\mathbf{f}_c^B(t)&=  \sum_{i=1}^{6} \mathbf{R}_{P_i}^B(\alpha(t))\mathbf{f}_{P_i}(t),
 \label{eq:f_c}\\
 \boldsymbol{\tau}_c^B(t)
 &=  \sum_{i=1}^{6} \mathbf{R}_{P_i}^B(\alpha(t)) \left(  \mathbf{p}_{P_i}^B \times  \mathbf{f}_{P_i}(t)+  \boldsymbol{\tau}_{P_i}^d(t)\right),\label{eq:tau_c} 
\end{align}
\endgroup
where the vector $\mathbf{p}_{P_i}^B \in \mathbb{R}^3$ identifies the position of the $i$-th propeller spinning center in $\mathscr{F}_B$, and the rotation matrix $\mathbf{R}_{P_i}^B(\alpha(t))  \in SO(3)$ describes the orientation of $\mathscr{F}_{P_i}$  with respect to (w.r.t.) $\mathscr{F}_B$. More specifically, it holds that
\begin{align}
&\mathbf{p}_{P_i}^B = \ell \,
\mathbf{R}_z \left({\tfrac{\pi}{6}+}(i-1) \tfrac{\pi}{3} \right) \mathbf{x}_B,\\
&\mathbf{R}_{P_i}^B(\alpha(t)) = \mathbf{R}_z \left({\tfrac{\pi}{6}+}(i-1) \tfrac{\pi}{3} \right)  \mathbf{R}_{x}( (-1)^i\, \alpha(t)),
\end{align}
where $\ell \in \mathbb{R}_{\geq0}$ denotes the distance between $O_B$ and any propeller spinning center, assumed to lie in the $(\mathbf{x}_B,\mathbf{y}_B)$ plane, and $\mathbf{R}_x(\cdot)$ and $\mathbf{R}_z(\cdot)$ are the canonical rotation matrices about the $x$ and $z$ axes, respectively.

According to the Newton-Euler approach, the dynamics of an $\alpha$-TingH is described by the following equations
\begin{align}
   \dot{\mathbf{p}}_B^W(t) &= \mathbf{v}_B^W(t) \label{eq::UAV_dynamic_iquation1} \\
   \dot{\mathbf{R}}_B^W(t) &= \mathbf{R}_B^W(t) [\boldsymbol{\omega}_B^B(t)]_\times \label{eq::UAV_dynamic_iquation2} \\
    m\, \ddot{\mathbf{p}}_B^W(t) &= m g\, \mathbf{e}_3 + \mathbf{R}_B^W(t)\left( \mathbf{f}_c^B(t) +\mathbf{f}_a^B(t) + \mathbf{f}_i^B(t) \right) \label{eq::UAV_dynamic_equation3}\\
    \mathbf{J}^B\, \dot{\boldsymbol{\omega}}_B^B(t) &= -\boldsymbol{\omega}_B^B(t) \times \mathbf{J}^B \boldsymbol{\omega}_B^B(t) +  \boldsymbol{\tau}_c^B(t) +\boldsymbol{\tau}_a^B(t)\label{eq::UAV_dynamic_equation4} 
\end{align}
where $\mathbf{p}_B^W(t) \in \mathbb{R}^3$ and $\mathbf{R}_B^W(t) \in SO(3)$ denote the position and orientation of $\mathscr{F}_B$ w.r.t. $\mathscr{F}_W$.
The vectors $\mathbf{v}_B^W(t) \in \mathbb{R}^3$ and $\boldsymbol{\omega}_B^B(t) \in \mathbb{R}^3$ represent the linear velocity of the $\alpha$-TingH expressed in $\mathscr{F}_W$ and its angular velocity expressed in $\mathscr{F}_B$, respectively, with $[\boldsymbol{\omega}_B^B(t)]_\times \in \mathbb{R}^{3 \times 3}$ denoting the associated skew-symmetric matrix.
The scalars $g \in \mathbb{R}_{\leq 0}$ and $m \in \mathbb{R}_{\geq 0}$ denote the gravitational acceleration and the total mass of the platform, while $\mathbf{J}^B \in \mathbb{R}^{3 \times 3}$ is the positive-definite inertia matrix of the vehicle, expressed in $\mathscr{F}_B$. To obtain a more realistic representation of the UAV dynamics, the principal aerodynamic phenomena influencing multi-rotor behavior at non-zero velocities (such as blade flapping, air friction, and dihedral effects) are incorporated into~\eqref{eq::UAV_dynamic_equation3}–\eqref{eq::UAV_dynamic_equation4} through the terms $\mathbf{f}_a^B(t), \boldsymbol{\tau}_a^B(t) \in \mathbb{R}^3$, in $\mathscr{F}_B$ and according to~\cite{abbaraju2021aerodynamic}.

In~\eqref{eq::UAV_dynamic_equation3}–\eqref{eq::UAV_dynamic_equation4}, we also model the presence of a rigid interaction tool, assumed to be a lightweight stick attached so as to be approximately aligned with the platform CoM. Under this assumption, the term $\mathbf{f}_i^B(t) \in \mathbb{R}^3$ represents the interaction forces applied at the vehicle CoM, while the interaction torque is considered negligible. Such an assumption is well justified in application scenarios involving quasi-static contact tasks (e.g., pushing or probing), where the interaction tool is typically short and aligned with the platform frame. We further assume that $\mathbf{f}_i^B(t)$ is known and is treated as an external disturbance, which will be subsequently compensated.

Finally, in modeling the dynamics of an $\alpha$-TingH, it is worth accounting also for the dynamics of the servomotors enabling propeller tilting, in line with the most common mechanical design solutions~\cite{kamel2018voliro}.
Their action is modeled via a first-order system approximation, yielding $\dot{\alpha}_{P_i}(t) = \frac{1}{\tau_\alpha}\left( \alpha_{P_i}^\ast(t) - \alpha_{P_i}(t) \right)$,
where $\alpha_{P_i}^\ast(t) \in  \Gamma_\alpha$ denotes the commanded tilt angle and $\tau_\alpha \in \mathbb{R}_{>0}$ is the servomotor time constant.
\begin{table}[!t]
    \renewcommand{\arraystretch}{1.05}
    \centering
      \caption{Parameters of the $\alpha$-TingH platform case study}
    \label{tab:platform_parameters}
    \begin{tabular}{ll|c}
       \toprule
        $m$ & $  [\si{\kilogram}]$ & $3.500$ \\
        $\mathbf{J}$ & $ [\si{\kilogram\meter\squared}]$ & $\; \mathrm{diag}\left(\colvec{0.147 \; 0.155 \; 0.251} \right)$ \\
        $\ell $ & $ [\si{\meter}]$ & $0.385$ \\
        $c_f $ & $ [\si{\newton/\hertz\squared}]$ & $1.500 \times 10^{-3}$ \\
        $c_\tau $ & $ [\si{\newton\meter/\hertz\squared}]$ & $4.590 \times 10^{-5}$ \\
        $\bar{\omega} $ & $ [\si{\hertz}]$ & $108.00$ \\
        $\tau_{\alpha} $ & $ [\si{\second}]$ & $5 \times 10^{-3}$ \\
        \bottomrule
    \end{tabular}
\end{table}

\begin{remark}
    Hereafter, the numerical results refer to a case study of the $\alpha$-TingH  characterized by the parameters in Table~\ref {tab:platform_parameters}. From now on, frames and time dependence are omitted when not essential.
   \end{remark}

\subsection{Zero-Moment Control Force Polytope}\label{sec:2:polytope}
\begin{figure}[!t]
    \centering
    \includegraphics[width=0.35\linewidth]{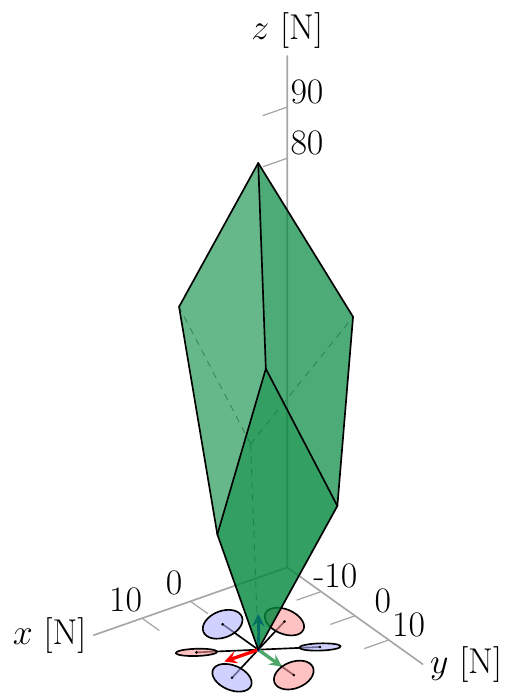}
    \caption{3D representation of the zero-moment control force polytope for $\alpha = 25^\circ$.}
    \label{fig:polytopeF_image}
\end{figure}

Recalling~\cite{perin2024star}, by introducing the {control input vector} $\mathbf{u} = \colvec{ \omega_{P_1}^2 \; \cdots \; \omega_{P_6}^2 }^\top \in \mathcal{U}$ with $ \mathcal{U} = \prod_{i=1}^6 [0,\bar{\omega}^2]$, the control force~\eqref{eq:f_c} and the control moment~\eqref{eq:tau_c}  can be expressed as $\mathbf{f}_{c}=\mathbf{F}_\alpha \mathbf{u}$ and $\boldsymbol{\tau}_{c}= \mathbf{M}_\alpha \mathbf{u}$,
where the matrices $\mathbf{F}_\alpha, \mathbf{M}_\alpha \in \mathbb{R}^{3 \times 6}$ depend on the (time-varying) tilt angle $\alpha$ and the block matrix $\mathbf{C}_\alpha = \colvec{\mathbf{F}_\alpha^\top ; \mathbf{M}_\alpha^\top}^\top \in \mathbb{R}^{6 \times 6}$ defines the $\alpha$-TingH control allocation matrix. The platform is fully actuated when $\mathbf{C}_\alpha$ is full rank ($\alpha \neq 0$), in which case the feasible control force space is three-dimensional.

Beyond full actuation, the platform is fully decoupled when translational forces can be generated independently of the control moments, allowing pure translational motions without affecting the rotational dynamics.
From a mathematical standpoint, the decoupling condition stands when 
$\mathbf{f}_c$ can be assigned in the \textit{control force space} 
$\mathcal{F}(\alpha)  \subseteq \text{Im}(\mathbf{F}_\alpha)$, independently of $\boldsymbol{\tau}_c$, i.e.,when the \textit{zero-moment control force space} $\mathcal{F}_0(\alpha)  \subseteq \text{Im}(\mathbf{F}_\alpha)$ coincides with $\mathcal{F}(\alpha)$. Formally, it holds that
\begin{align}
\mathcal{F}(\alpha) &= \left\{ \mathbf{f}_c \in \mathbb{R}^3 \middle|
\mathbf{f}_c = \mathbf{F}_\alpha \mathbf{u}, \mathbf{u} \in \mathcal{U} \right\}, \\
\mathcal{F}_0(\alpha) &= \left\{ \mathbf{f}_c \in \mathbb{R}^3 \middle|
\mathbf{f}_c = \mathbf{F}_\alpha \mathbf{u}, \mathbf{u} \in \left(\mathcal{U}\cap \ker(\mathbf{M}_\alpha)\right) \right\}.\label{eq:Fspace_def_2}
\end{align}

In~\cite{perin2024star}, it is shown that, for any fixed tilt angle, the expression~\eqref{eq:Fspace_def_2} of $\mathcal{F}_0(\alpha)$ corresponds to a bounded, convex, finite polytope that can be represented and analyzed geometrically (see an example in Figure~\ref{fig:polytopeF_image}). In particular, its geometric characteristics (such as volume) can thus be interpreted as indices of the platform’s maneuverability.

In the case of $\alpha$-TingH platforms, the time variation of the tilt angle plays a crucial role in shaping the zero-moment control force space, as it continuously modifies the geometry of the associated force polytope. Under the stated assumption for the interaction tool, this polytope directly defines the range of feasible interaction forces that can be applied without inducing rotational effects on the platform. Consequently, different values of $\alpha$ lead to distinct actuation and interaction capabilities, making the angle selection a key control aspect.


\section{{Control Framework Overview}}\label{sec:my_method}
\begin{figure}
    \centering
    \includegraphics[width=0.9\columnwidth]{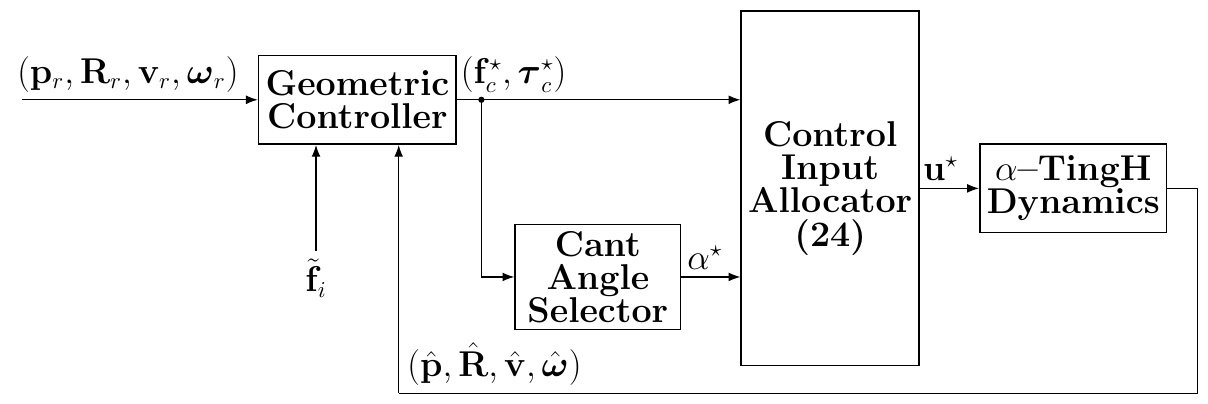}
    \caption{Overall control architecture of the proposed framework. The geometric full-pose controller generates the desired wrench, while an offline-computed LUT is exploited to select a feasible and efficient tilt angle $\alpha$ prior to control allocation.}
    \label{fig:proposed_controller}
\end{figure}

This section presents the proposed trajectory-tracking control architecture for $\alpha$-TingH platforms, shown in Figure~\ref{fig:proposed_controller}, consisting of a geometric full-pose controller, a cant-angle selector, and a standard control allocator.

\subsection{Geometric Pose Controller}
Described in~\cite{ryll2021fast}, the adopted state-of-the-art controller leverages the  geometry of the $SE(3)=\mathbb{R}^3 \times SO(3)$ manifold to achieve pose tracking.
This approach is particularly suitable for fully actuated platforms, as it enables direct control of both position and orientation while preserving robustness to nonlinear dynamics.

For $\alpha$-TingH UAVs, the nonlinear motion equations~\eqref{eq::UAV_dynamic_equation3}--\eqref{eq::UAV_dynamic_equation4} can be rewritten in a compact form suitable for feedback linearization as
\begin{align}
   \colvec{\ddot{\mathbf{p}} \\ \dot{\boldsymbol{\omega}}} &= \colvec{ g\, \mathbf{e}_3 +  \frac{1}{m}\mathbf{R}\left(\mathbf{f}_a(\mathbf{v},\alpha) + \mathbf{f}_i \right) \\ 
    \mathbf{J}^{-1}\left(-\boldsymbol{\omega} \times \mathbf{J} \boldsymbol{\omega} +\boldsymbol{\tau}_a(\boldsymbol{\omega},\alpha) \right)}+\colvec{  \frac{1}{m}\mathbf{R} & \mathbf{0} \\
        \mathbf{0} & \mathbf{J}^{-1}}
        \colvec{
            \mathbf{f}_c \\ \boldsymbol{\tau}_c
        }\\
&=\colvec{\mathbf{f}_e(\mathbf{R},\mathbf{v}, \alpha, \mathbf{f}_i)\\\boldsymbol{\tau}_e(\boldsymbol{\omega},\alpha)}+\colvec{  \frac{1}{m}\mathbf{R} & \mathbf{0} \\
        \mathbf{0} & \mathbf{J}^{-1}}  \colvec{
            \mathbf{f}_c \\ \boldsymbol{\tau}_c
        }
  \label{eq:gc_dyn1}
\end{align}
where the dependence of $\mathbf{f}_a$ and $\boldsymbol{\tau}_a$ on the platform linear and angular velocities, and on the tilt angle, is made explicit.

Hereafter, the UAV state is assumed to be feedback by a suitable estimator/measurement system, yielding $(\hat{\mathbf{p}},\hat{\mathbf{R}},\hat{\mathbf{v}},\hat{\boldsymbol{\omega}}) \in SE(3) \times \mathbb{R}^6 $. The tilt angle $\alpha$ is treated as an internal variable of the control framework, whereas the interaction force $\mathbf{f}_i$ is either directly measured or indirectly estimated, and a close approximation $\tilde{\mathbf{f}}_i \in \mathbb{R}^3$ is assumed to be available.

Given a reference pose trajectory $(\mathbf{p}_r,\mathbf{R}_r) \in SE(3)$, with corresponding reference velocities $(\mathbf{v}_r,\boldsymbol{\omega}_r) \in \mathbb{R}^6$, the position, orientation, linear velocity, and angular velocity errors are defined as 
\begingroup
\setlength{\abovedisplayskip}{4pt}
\setlength{\belowdisplayskip}{4pt}
\setlength{\abovedisplayshortskip}{4pt}
\setlength{\belowdisplayshortskip}{4pt}
\begin{align}
&\mathbf{e}_p = \hat{\mathbf{p}} - \mathbf{p}_r \label{eq:pos_error}\\
&\mathbf{e}_R = \frac{1}{2}\left( \mathbf{R}_r^\top \hat{\mathbf{R}} - \hat{\mathbf{R}}^\top \mathbf{R}_r \right)^\vee \label{eq:geodesic_error_SO3}\\
& \mathbf{e}_v = {\hat{\mathbf{v}}} - {\mathbf{v}}_r\\
& \mathbf{e}_\omega = \hat{\boldsymbol{\omega}} - \hat{\mathbf{R}}^\top \mathbf{R}_r \boldsymbol{\omega}_r
\end{align}
\endgroup
where $(\cdot)^\vee$ denotes the vee operator, i.e., the inverse of the $[\cdot]_\times$ mapping. Then, the desired control wrench can be explicitly computed as 
\begin{equation}
 \colvec{
            \mathbf{f}_c^\ast \\ \boldsymbol{\tau}_c^\ast
        } = \colvec{  \frac{1}{m}\hat{\mathbf{R}} & \mathbf{0} \\
        \mathbf{0} & \mathbf{J}^{-1}}^{-1}\left(-\colvec{\mathbf{f}_e(\hat{\mathbf{R}},\hat{\mathbf{v}}, \alpha, \tilde{\mathbf{f}}_i)\\\boldsymbol{\tau}_e(\hat{\boldsymbol{\omega}},\alpha)} + \colvec{\mathbf{w}_p \\ \mathbf{w}_o}\right),
\label{eq:gc_control_wrench}
\end{equation}
where $\mathbf{w}_p,  \mathbf{w}_o \in \mathbb{R}^3$ are virtual control inputs enforcing the desired system behavior.
In detail, a linear feedback law is adopted to define the virtual control inputs $\mathbf{w}_p$ and $\mathbf{w}_o$, namely
\begin{align}
    \mathbf{w}_p &= \ddot{\mathbf{p}}_r
    - \mathbf{K}_{pd} {\mathbf{e}}_v  - \mathbf{K}_{pp} \mathbf{e}_p
    - \mathbf{K}_{pi} \int_{0}^{t} \mathbf{e}_p(\tau)\, d\tau, \label{eq:vp_law}\\
    \mathbf{w}_o &= \dot{\boldsymbol{\omega}}_r
    - \mathbf{K}_{od} \mathbf{e}_\omega  - \mathbf{K}_{op} \mathbf{e}_R
    - \mathbf{K}_{oi} \int_{0}^{t} \mathbf{e}_R(\tau)\, d\tau, \label{eq:vR_law} 
\end{align}
where $\mathbf{K}_{pp}, \mathbf{K}_{pd}, \mathbf{K}_{pi}, \mathbf{K}_{op}, \mathbf{K}_{od}, \mathbf{K}_{oi} \in \mathbb{R}^{3 \times 3}$ are diagonal, positive definite gain matrices.

\subsection{Cant--Angle Selector}\label{sec:cant_angle_selector}
Given the desired control wrench, the optimal angle $\alpha^\ast$ is determined by a force--based cant--angle selector. Focusing only on the force component of the desired wrench is motivated by the typical requirements of interactive tasks performed by platforms equipped with a rigid interaction tool, as described in Section~\ref{sec:model_and_polytope}.

The selector consists of:
$i)$ an \textit{offline-computed LUT} associating discrete tilt angles in $\Gamma_\alpha$ with the corresponding zero-moment control force polytopes;
$ii)$ an \textit{online candidate tilt angles identification} step grounded on a safety-margin containment condition for the desired control force; and
$iii)$ an \textit{online optimization procedure} selecting the optimal $\alpha^\ast \in \Gamma_\alpha$ among the admissible candidate angles.

\subsubsection{Offline-Computed LUT}\label{subsec:offline_LUT}
The LUT underlying the proposed cant-angle selector stores the boundaries of the zero-moment control force polytope for discretized values of the cant angle $\alpha \in \Gamma_\alpha$, using a fixed increment $\Delta\alpha$. From a practical standpoint, each LUT entry is computed using the results in~\cite{perin2024star}. 

Specifically, for any
$\mathbf{u} \in \mathcal{U} \cap \ker(\mathbf{M}_\alpha)$, it holds that
$u_k = u_{k+3} = \tilde{u}_k$, with $\tilde{u}_k \in [0,\bar{\omega}^2]$ and
$k \in \{1,2,3\}$. 
In other words, opposite propellers w.r.t. the platform CoM are required to spin at the same rate to generate a zero-moment control force in $\mathcal{F}_0(\alpha)$.
 Intuitively, this property follows from the balanced CW/CCW propeller spinning directions, alongside the alternating tilt pattern between adjacent rotors.

In light of this fact, by defining $\tilde{\mathbf{u}} = \colvec{\tilde{u}_{1} \; \tilde{u}_{2} \; \tilde{u}_{3}}^\top$, any vector in the zero-moment control force space can be expressed as
\begin{equation}
    \mathbf{f}_c = 2c_f \mathbf{F}_\alpha \mathbf{S} \tilde{\mathbf{u}}  \quad \text{with} \quad \mathbf{S}=\colvec{\mathbf{I}_{3\times 3}\\ \mathbf{0}_{3 \times 3}}. \label{eq:force_param}
\end{equation}
Hence, the vertices of the polytope $\mathcal{F}_0(\alpha)$ are obtained from~\eqref{eq:force_param} by evaluating the components of $\tilde{\mathbf{u}}$
 at the extrema of the interval $[0,\bar{\omega}^2]$. 
The convex hull of the resulting force vectors defines the boundary of $\mathcal{F}_0(\alpha)$.

The collection of polytopes stored in the LUT provides a compact representation of the force-generation capabilities of the $\alpha$-TingH platform and significantly reduces the computational burden at runtime.

\subsubsection{Online Candidate Tilt Angles Identification}\label{subsec:force_feasibility_analysis}
The candidate tilt-angle identification step determines the set of (discretized) cant angles for which the platform can generate forces in all directions around the desired control force vector within a prescribed safety margin.
Geometrically, this requirement is expressed by imposing that a sphere with radius $r \in \mathbb{R}_{\geq 0}$, centered at $\mathbf{f}_c^\ast$, be fully contained within the zero-moment force polytope associated with each candidate $\alpha$. 

Practically, the set $\mathcal{A}$ is obtained by scanning the LUT and retaining the values of $\alpha$ for which
$
\mathcal{B}(\mathbf{f}_c^\ast, r) \subseteq \mathcal{F}_0(\alpha),
$
where $\mathcal{B}(\mathbf{f}_c^\ast, r)$ denotes the Euclidean ball centered on $\mathbf{f}_c^\ast$ with radius $r$. The radius $r$ is defined as the product of a nominal robustness margin $r^\ast \in \mathbb{R}_{\geq 0}$ and a relaxation parameter $c_r \in (0,1]$ such that, as $c_r \to 0$, the admissible set $\mathcal{A}$ includes tilt angles for which the desired control force $\mathbf{f}_c^\ast$ lies arbitrarily close to the boundary of the zero-moment force polytope.

The identification of $\mathcal{A}$ is first performed with $c_r = 1$ (nominal robustness). 
If no $\alpha$ satisfies this condition, the procedure is repeated once with an assigned $c_r \in (0,1)$ to relax the robustness requirement. 
If no tilt angle satisfies the relaxed condition, the desired control force is declared unfeasible, keeping the cant angle at its previous value to preserve configuration continuity, and raising an error condition.

\subsubsection{Online Optimization Procedure}\label{sec:5}
The optimal tilt angle $\alpha^\ast$ is then selected from the set $\mathcal{A}$ to ensure both system energy efficiency and smooth behavior.  
This is achieved by solving the optimization problem  
\begin{equation}
    \alpha^\ast = \alpha(t^+) = \text{argmin}_{\alpha(t^+) \in \mathcal{A}} J(\alpha(t^+),\alpha(t)),
\end{equation}
where $\alpha(t)$ is the current cant angle, and the cost function $J(\alpha(t^+),\alpha(t))$ is defined as  
\begin{equation}
    J(\alpha(t^+),\alpha(t)) = c_1 \, \lvert \alpha(t^+) \rvert + c_2 \, \lvert \alpha(t^+) - \alpha(t) \rvert,
    \label{eq:cost_function}
\end{equation}
with tunable coefficients $c_1,c_2 \in \mathbb{R}_{\geq 0}$.  
The first term in~\eqref{eq:cost_function} penalizes deviation from the underactuated configuration ($\alpha=0$), which typically corresponds to a more energy-efficient regime, while the second term penalizes large changes relative to the current tilt angle, promoting temporal smoothness and avoiding abrupt transitions.

\subsection{Control Allocator}
Once the optimal tilt angle $\alpha^\ast$ is identified, the mapping between the control input vector and the desired control wrench is defined by the corresponding control allocation matrix $\mathbf{C}_{\alpha^\ast}$. The optimal control input vector $\mathbf{u}^\ast$ can then be computed as
\begin{equation}
    \mathbf{u}^\ast
    =
    \mathbf{C}_{\alpha^\ast}^{\dagger}
    \colvec{\mathbf{f}_c^\ast \\ \boldsymbol{\tau}_c^\ast},
\end{equation}
where the pseudoinverse $\mathbf{C}_{\alpha^\ast}^{\dagger}
=
\mathbf{C}_{\alpha^\ast}^\top
\left(
\mathbf{C}_{\alpha^\ast}^\top
\mathbf{C}_{\alpha^\ast}
\right)^{-1} \in \mathbb{R}^{6 \times 6}$
 coincides with the inverse of $\mathbf{C}_{\alpha^\ast}$ for any $\alpha^\ast \neq 0$.


\section{Assessment and Validation}\label{sec:validation}
This section presents the results of the assessment and validation process of the proposed control framework conducted in MATLAB/Simulink environment.

\subsection{Simulation Scenario}\label{subsec:simulation_scenario}
\begin{table}[!t]
    \renewcommand{\arraystretch}{1.1}
    \centering
    \caption{Model Parameters of MoCap and Force Sensor}
    \label{tab:covariances_bias}
    \footnotesize
    \begin{tabular}{cll|l}
        \toprule
        \multirow{4}{*}{MoCap}
        & $\boldsymbol{\Sigma}_p$       & [$\si{\meter\squared}$]                           & $10^{-7} \cdot \; \mathrm{diag}\left(\colvec{4.099 \; 2.838 \; 0.211} \right)$ \\
        & $\boldsymbol{\Sigma}_R$       & [$\si{\deg\squared}$]                             & $\mathrm{diag}\left(\colvec{0.0012 \; 0.0011 \; 0.0011} \right)$ \\
        & $\boldsymbol{\Sigma}_v$       & [m$^2$/s$^2$] & $10^{-6} \cdot \; \mathrm{diag}\left(\colvec{2.050 \; 1.419 \; 0.105} \right)$ \\
        & $\boldsymbol{\Sigma}_\omega$  & [deg$^2$/s$^2$] & $\; \mathrm{diag}\left(\colvec{0.0024 \; 0.0022 \; 0.0022} \right)$ \\
        \midrule
        \multirow{2}{*}{Force Sensor}
        & $\boldsymbol{\Sigma}_f$       & [$\si{\newton\squared}$]    & $10^{-3} \cdot \; \mathrm{diag}\left(\colvec{2.5 \; 2.5 \; 2.5} \right)$ \\
        & $\mathbf{m}_f$       & [$\si{\newton}$]    & $\left[ 0.1 \; 0.1 \; 0.1 \right]^\top$ \\
        \bottomrule
    \end{tabular}
\end{table}

The simulation scenario emulates in-contact tasks characterized by a time-varying interaction force. 
Initially, the $\alpha$-TingH takes off and reaches a stationary hovering condition 
$(\mathbf{p}_r,\mathbf{R}_r)=\left(\colvec{0 \; 0 \; 1}^\top,\mathbf{I}_3\right)$ within $20\si{\second}$. 
Subsequently, the vehicle is subjected to an external force applied at its CoM, consistent with the interaction model described in Section~\ref{sec:model_and_polytope}. 
Both the magnitude and direction of the applied force $\mathbf{f}_i$ vary over time, mimicking disturbances experienced during the task.

The interaction force is assumed to be measured by a force sensor operating at $100\si{\hertz}$. 
The measurement is modeled as $\tilde{\mathbf{f}}_i=\mathbf{f}_i+\mathbf{n}_f$, where $\mathbf{n}_f$ is Gaussian noise with mean $\mathbf{m}_f\in\mathbb{R}^3$ and covariance $\boldsymbol{\Sigma}_f\in\mathbb{R}^{3\times3}$ (Table~\ref{tab:covariances_bias}).

The $\alpha$-TingH state is assumed to be retrieved through an external motion capture (MoCap) system at $100\si{\hertz}$. 
The impact of the MoCap system is modeled by introducing a $12\si{\milli\second}$ delay and additive zero-mean Gaussian noise affecting position, orientation, linear velocity, and angular velocity with covariances $\boldsymbol{\Sigma}_p$, $\boldsymbol{\Sigma}_R$, $\boldsymbol{\Sigma}_v$, and $\boldsymbol{\Sigma}_\omega$ (Table~\ref{tab:covariances_bias}).

Finally, the offline LUT storing the force polytope boundaries is constructed by discretizing the tilt angle $\alpha\in\Gamma_\alpha$ with resolution $\Delta\alpha=1\si{\degree}$.
\begin{figure*}[t!]
    \centering
\begin{align}
    (\alpha^\ast,\mathbf{u}^\ast) = (\alpha(t^+),\mathbf{u}(t^+)) =\text{argmin}_{\substack{\alpha(t^+) \in \Gamma_\alpha\\\mathbf{u}(t^+) \in \bar{\mathcal{U}}}} \left( \left\Vert \mathbf{C}_{\alpha(t^+)} \mathbf{u}(t^+) - \colvec{\mathbf{f}_c^\ast \\ \boldsymbol{\tau}_c^\ast} \right\Vert^2 + J(\alpha(t^+),\alpha(t))\right) 
    \label{eq:baseline}
\end{align}
\end{figure*}

\subsection{Effects of Different Weights}\label{subsec:weights}
Since the cant--angle selection relies on the weighted cost function~\eqref{eq:cost_function}, we investigate the $\alpha$–TingH behavior under different $c_1/c_2$ ratios.
\begin{table}[!t]
    \renewcommand{\arraystretch}{1.1}
    \centering
      \caption{Numerical Simulation Parameters: Controller Gains, Cant-Angle Selection, and Force Profile}
    \label{tab:controller_gains_and_co}
    \begin{tabular}{lc|lc}
       \toprule
        $\mathbf{K}_{pp}$ & $\mathrm{diag}\left( \left[ 30 \; 30 \; 70 \right] \right)$ &  $\mathbf{K}_{op}$ & $\mathrm{diag}\left( \left[ 20 \; 20 \; 5 \right] \right)$ \\
        $\mathbf{K}_{pd}$ & $\mathrm{diag}\left( \left[ 10 \; 10 \; 10 \right] \right)$ &  $\mathbf{K}_{od}$ & $\mathrm{diag}\left( \left[ 10 \; 10 \; 10 \right] \right)$ \\
        $\mathbf{K}_{pi}$ & $\mathrm{diag}\left( \left[ 30 \; 30 \; 40 \right] \right)$ &         $\mathbf{K}_{oi}$ & $\mathrm{diag}\left( \left[ 0.1 \; 0.1 \; 1 \right] \right)$ \\
        \midrule
        $r^\ast \left[\si{\newton}\right]$ & $1$ &
        $c_r \left[-\right]$ & $\frac{1}{3}$\\
        \midrule
        $\mathcal{T}_w$ [s] & \multicolumn{3}{c}{$\{ 0, 20, 38, 60, 80\}$} \\
        $\mathcal{F}_w$ [N] &  \multicolumn{3}{c}{$\left\{ \colvec{0 \\0\\ 0}, \colvec{0 \\0\\ 0}, \colvec{0 \\-8\\ 10}, \colvec{-10 \\7\\ 2}, \colvec{-6 \\-8\\ 0}\right\}$}\\
        \bottomrule
    \end{tabular}
\end{table}
\begin{figure}
    \centering
   \includegraphics[width=0.8\linewidth]{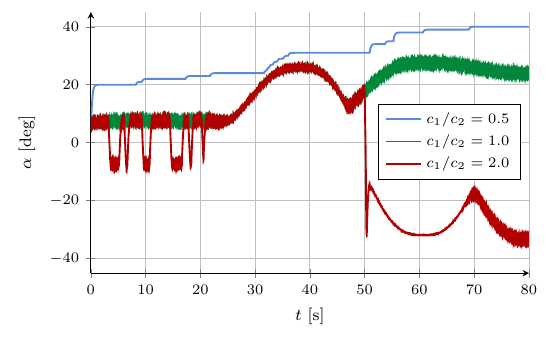}
    \caption{Validation of the influence of the weight coefficients $c_1$ and $c_2$ on the cant-angle selection, with identical force profile and radius $r^\ast$.}
    \label{fig:different_weights}
\end{figure}
\begin{figure*}[!t]
    \centering
    \includegraphics[width=0.65\linewidth]{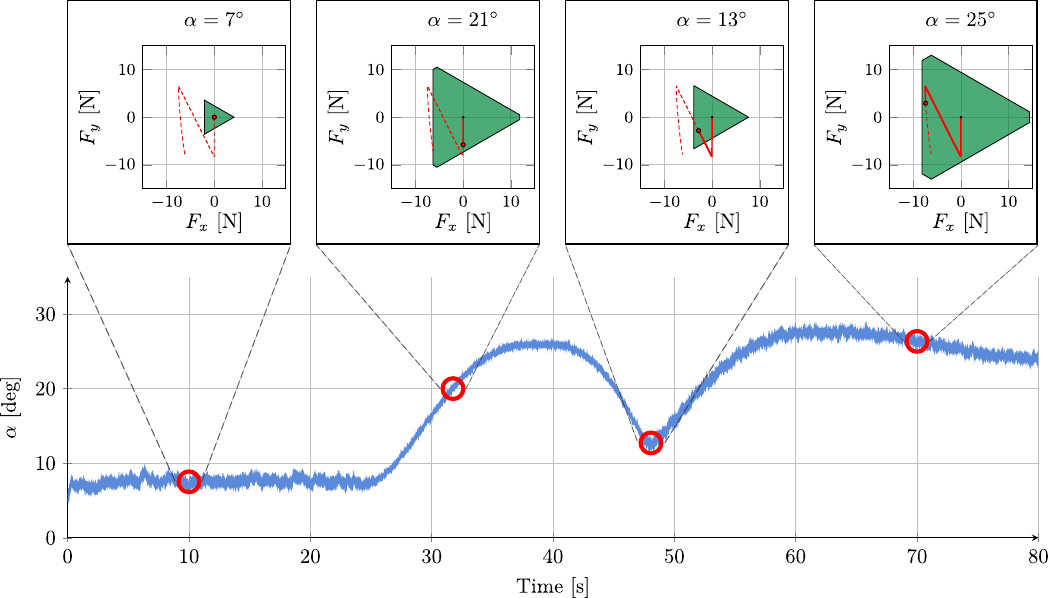}
    \caption{Cant-angle trajectory $\alpha(t)$ and corresponding zero-moment force polytope sections at four selected instants on the hovering plane, together with the applied (red dashed) and instantaneous required (red marker) forces, confirming that the geometric containment condition is satisfied throughout the maneuver.}
    \label{fig:alpha_and_sections}
\end{figure*}

In this sensitivity analysis, the nominal interaction force profile is generated from a sequence of waypoints $\left(\mathcal{T}_w,\mathcal{F}_w\right)$ and interpolated with cubic polynomials.
The adopted waypoints, along with the controller gains and the candidate angle identification parameters, are reported in Table~\ref{tab:controller_gains_and_co}.

The ratio $c_1/c_2$ is varied over the set $\{ 0.5,\,1.0,\,2.0 \}$, and the resulting trends are reported in Figure~\ref{fig:different_weights}. The following observations can be made.
For $c_1/c_2=0.5$, the smoothness term dominates the cost function and $\alpha(t)$ evolves conservatively, with small incremental variations and limited sensitivity to efficiency changes. 
When $c_1/c_2=1.0$, the two terms contribute equally, leading to a balanced behavior that avoids both excessive oscillations and overly conservative responses. 
For $c_1/c_2=2.0$, the efficiency term prevails, resulting in more aggressive variations of $\alpha(t)$ and occasional abrupt transitions (e.g., around $50\si{\second}$).

Based on this analysis, $c_1/c_2 = 1.0$ is selected for the Monte Carlo validation and physics-based simulation discussed in the following, as it is the most balanced ratio among the considered set.

\subsection{Baseline Comparison}\label{subsec:MC_analysis}
The performance of the proposed control framework is compared with a baseline solution in which the cant angle $\alpha$ and the control input vector $\mathbf{u}$ are jointly computed online by solving a constrained nonlinear optimization problem. In more detail, the selected baseline controller is an alternative to the combination of a cant--angle selector and a control allocator, and it outputs the pair $(\alpha^\ast,\mathbf{u}^\ast)$ as in~\eqref{eq:baseline}.
In practice, the nonlinear program~\eqref{eq:baseline} is solved using a SQP-method, in line with joint optimization strategies commonly adopted for tilting multirotor platforms (see, e.g.,~\cite{ruggia2025momav}). 

For comparative analysis, the proposed control framework is evaluated for different values of the nominal robustness margin, namely $r^\ast \in \{0.5,1,3,5\}\si{\newton}$. 
For each value, $100$ Monte Carlo (MC) runs are performed with fixed parameters $c_r=\frac{1}{3}$ and $c_1=c_2=\frac{1}{2}$ over $N\in\mathbb{N}$ simulation steps. 
In each run, the nominal interaction force profile $\mathbf{f}_i(t)$ is generated from waypoint samples at times $\mathcal{T}_w$ (Table~\ref{tab:controller_gains_and_co}), where the $x$ and $y$ components are drawn independently from $\mathcal{U}([-15,15])\,\si{\newton}$ and the $z$ component from $\mathcal{U}([0,15])\,\si{\newton}$, consistent with the pushing/probing scenario. 
The resulting discrete samples are then interpolated using cubic polynomials to obtain a continuous-time force profile $\mathbf{f}_i(t)$.

Figure~\ref{fig:alpha_and_sections} provides insight into the functioning of the cant-angle selector during a MC run with $r^\ast = 1.0\si{\newton}$. For selected time instants, it shows the profile of $\alpha(t)$ and the corresponding intersections of the zero-moment force polytope with the hovering plane $z = mg$ (green areas).  
This visualization illustrates how the cant-angle selection mechanism dynamically adapts the tilting configuration of the $\alpha$--TingH platform in response to the applied interaction force (red dotted/solid line).

\begingroup
\setlength{\abovedisplayskip}{4pt}
\setlength{\belowdisplayskip}{4pt}
\setlength{\abovedisplayshortskip}{4pt}
\setlength{\belowdisplayshortskip}{4pt}
Several key performance indicators (KPIs) are evaluated.
\begin{itemize}
    \item Both position and orientation errors~\eqref{eq:pos_error}--\eqref{eq:geodesic_error_SO3} are analyzed to assess controller accuracy by evaluating the norm of the mean error vector and the corresponding RMS values, defined as
\begin{align}
    &\|\bar{\mathbf e}_\bullet\| = \left\lVert \frac{1}{N} \sum_{t=1}^{N} \mathbf e_{\bullet}(t) \right\rVert, \; \; \mathrm{RMS}_{\mathbf{e}_\bullet} = \sqrt{\frac{1}{N} \sum_{t=1}^{N} \|\mathbf e_{\bullet}(t)\|^2},
\end{align}
where $(\cdot)_\bullet \in \{p,R\}$ denotes either the position or orientation components.
    \item The normalized Motor Effort Index (MEI) is investigated to quantify how intensely the actuation system is used relative to its saturation limit. In particular, we account for the average MEI, defined as
\begin{align}
\overline{\mathrm{MEI}} =
\frac{1}{N} \sum_{t=1}^{N}
\left(\frac{1}{\bar{\omega}} \max_{P_i} \, \omega_{P_i}(t)\right).
\end{align}
    \item The RMS value of the vector $\mathbf{u}$ is evaluated to quantify the overall control input effort:
\begin{equation}
    \mathrm{RMS}_\mathbf{u} = \sqrt{\frac{1}{N} \sum_{t=1}^{N} \|\mathbf{u}(t)\|^2}.
\end{equation}
    \item The Force Efficiency Index (FEI) quantifies how effectively the individual rotor contributions combine to produce the total control force.  We focus on its average value
\begin{equation}
\overline{\mathrm{FEI}} =
\frac{1}{N} \sum_{t=1}^{N}
\left(
\frac{1}{6 c_f \bar{\omega}^2} \left\| \sum_{i=1}^{6} \mathbf{f}_{P_i}(t) \right\|
\right).
\end{equation} 
    \item The mean computation time $\bar{t}_{c}$ is evaluated as the average allocation time per control step, measured from the desired control wrench to the resulting $\mathbf{u}$ and $\alpha$, and averaged over the simulation horizon and all MC runs.
\end{itemize}
\endgroup

\begin{table}[!t]
    \renewcommand{\arraystretch}{1.1}
    \centering
    \caption{$\mathrm{MC}$ KPIs for the baseline and proposed controller.}
    \label{tab:kpis}
\resizebox{1\columnwidth}{!}{
    \begin{tabular}{p{0.7cm}
                    >{\centering\arraybackslash}p{0.5cm}|
                    ccccc}
        \toprule
         & & \textbf{baseline} & \multicolumn{4}{c}{\textbf{proposed controller}}\\
         &  & & $r^\ast = 0.5$N & $r^\ast = 1$N & $r^\ast = 3$N & $r^\ast = 5$N \\
        \midrule
        $\bar{t}_c$            & $\left[ \si{\milli\second} \right]$       & $0.5792$  & $0.0565$    & $0.0545$    & $0.0536$     & $\mathbf{0.0521}$ \\
        $\|\bar{\mathbf{e}}_p\|$      & $\left[ \si{\meter} \right]$        & $0.0422$  & $\mathbf{0.0192}$    & $\mathbf{0.0192}$    & $0.0202$     & $0.0223$ \\
        $\mathrm{RMS}_{\mathbf{e}_p}$       & $\left[ \si{\meter} \right]$        & $0.0352$  & $\mathbf{0.0282}$    & $0.0280$    & $0.0291$     & $0.0313$ \\
        $\|\bar{\mathbf{e}}_R\|$      & $\left[ \si{\radian} \right]$      & $0.0028$  & $\mathbf{0.0004}$    & $0.0005$    & $0.0008$     & $0.0014$ \\
        $\mathrm{RMS}_{\mathbf{e}_R}$       & $\left[ \si{\radian} \right]$      & $0.0046$  & $\mathbf{0.0005}$    & $\mathbf{0.0005}$    & $0.0009$     & $0.0013$ \\
        $\overline{\mathrm{FEI}}$ & $\left[ - \right]$      & $0.6366$   & $\mathbf{0.6465}$      & $0.5523$      & $0.5355$       & $0.5212$ \\
        $\overline{\mathrm{MEI}}$ & $\left[ - \right]$      & $0.6532$    & $\mathbf{0.5238}$      & $0.5435$      & $0.5174$       & $0.5586$ \\
        $\mathrm{RMS}_\mathbf{u}$        & $\left[ \si{\hertz\squared} \right]$ & $5023$    & $\mathbf{4642}$      & $4649$      & $4864$       & $4865$ \\
        \bottomrule
    \end{tabular}}
\end{table}

Table~\ref{tab:kpis} summarizes the performance comparison between the baseline and the proposed control framework.
From a computational perspective, for all values of $r^\ast$, the proposed control framework achieves a substantial reduction in mean computation time $\bar{t}_c$ compared to the baseline.  
This confirms that the offline LUT effectively alleviates part of the online computational burden and makes the framework suitable for resource-constrained platforms.

Focusing on pose-tracking performance, a clear trend emerges when jointly analyzing translational and rotational errors.  
Selecting $r^\ast \in \{0.5, 1\}$N consistently provides the best overall pose accuracy, yielding the lowest mean and RMS errors for both position and orientation, with particularly pronounced improvements in the orientation-related metrics.
However, as $r^\ast$ increases, a gradual degradation in both translational and rotational accuracy is observed.

Furthermore, the proposed control framework achieves the highest $\overline{\mathrm{FEI}}$ when $r^\ast = 0.5$N, indicating a more constructive combination of individual rotor thrust vectors and reduced internal force cancellation, although it remains comparable to the baseline method.  
In contrast, lower efficiency is observed for $r^\ast \in \{1, 3, 5\}$N, confirming that larger radii tend to favor configurations that are feasible but suboptimal in terms of force generation.

Finally, the analysis of actuation effort highlights an important trade-off.  
The lowest $\mathrm{RMS}_\mathbf{u}$ is observed for $r^\ast = 0.5$N, while the baseline exhibits the highest value.  
This demonstrates that the proposed control framework can simultaneously reduce computational burden and maintain lower actuator usage.  
However, the non-monotonic behavior of the index indicates that increasing the sphere radius does not systematically decrease or increase actuator loading.  

Overall, the validation results indicate that selecting $r^\ast \in \{0.5, 1\}$N provides the most effective balance among tracking accuracy, computational cost, and actuation efficiency, whereas larger robustness margin leads to suboptimal tilt configuration, resulting in poorer force efficiency and pose--tracking performance.

\subsection{Complete Interaction Task Simulation}\label{subsec:full_task}

As a final validation step, the proposed framework is evaluated in a physics-based simulation environment. 
The simulation is implemented in \emph{Simscape} using the MATLAB/Simulink toolbox \emph{RotorSuite}~\cite{cigarini2026rotorsuite}, which provides a modular representation of the multirotor platform, explicitly modeling masses, inertias, and actuator dynamics of the system.

In the considered scenario, the $\alpha$--TingH platform takes off and navigates toward a vertical wall located at $x_w = 6\si{\meter}$ in $\mathscr{F}_W$. 
After approaching the wall, the vehicle performs three sequential interaction contacts at the points
$\{(6,0,1), (6,0,2), (6,0,3)\}$, each lasting $5\si{\second}$, emulating a surface inspection task.
The contact is modeled through the \textit{Spatial Contact Force} block in Simscape, representing the wall as a spring-damper system with stiffness $K_e = 10^6~\si{\newton/\meter}$ and damping $K_d = 10^3~\si{\newton\second/\meter}$. 
When the end-effector tip reaches the wall, the resulting reaction force acts along the wall normal direction and is applied at the vehicle CoM, consistently with the assumed alignment of the tool with the body-frame $x$-axis.
To maintain contact, the position reference offset along the $x$ direction is set $2\,\si{\centi\meter}$ beyond the wall surface.
\begin{table}[!t]
    \renewcommand{\arraystretch}{1.1}
    \centering
    \caption{Simscape Physics-Based Simulation: Parameters, \\Controller Gains, Cant-Angle Selection, and KPIs.}
    \label{tab:simscape_kpis}
    \begin{tabular}{p{1.2cm}c|p{1.2cm}c}
        \toprule
        \multicolumn{2}{r}{$m \left[\si{\kilogram}\right]$} &
        \multicolumn{2}{l}{$3.8$} \\
        \multicolumn{2}{r}{$\mathbf{J} \left[\si{\kilogram\meter\squared}\right]$} &
        \multicolumn{2}{l}{$\mathrm{diag}\left([0.107 \; 0.103 \; 0.205]\right)$} \\
        \midrule
        $\mathbf{K}_{pp}$ & $\mathrm{diag}\left( \left[ 6 \; 6 \; 15 \right] \right)$ &  $\mathbf{K}_{op}$ & $\mathrm{diag}\left( \left[ 10 \; 10 \; 10 \right] \right)$ \\
        $\mathbf{K}_{pd}$ & $\mathrm{diag}\left( \left[ 10 \; 10 \; 20 \right] \right)$ &  $\mathbf{K}_{od}$ & $\mathrm{diag}\left( \left[ 1 \; 2 \; 2 \right] \right)$ \\
        $\mathbf{K}_{pi}$ & $\mathrm{diag}\left( \left[ 1 \; 1 \; 1 \right] \right)$ &  $\mathbf{K}_{oi}$ & $\mathrm{diag}\left( \left[ 0.1 \; 0.01 \; 0.1 \right] \right)$ \\
        \midrule
        $r^\ast \left[\si{\newton}\right]$ & $1$ &
        $c_r \left[-\right]$ & $\frac{1}{3}$\\
        \midrule
        $\bar{t}_c \; [\si{\milli\second}]$ & $0.0798$ &
        $\|\bar{\mathbf{e}}_p\| \; [\si{\meter}]$ & $0.1340$ \\

        $\mathrm{RMS}_\mathbf{u} \; [\si{\hertz\squared}]$ & $4237$ &
        $\mathrm{RMS}_{\mathbf{e}_p} \; [\si{\meter}]$ & $0.1623$ \\

        $\overline{\mathrm{FEI}} \; [-]$ & $0.9131$ &
        $\|\bar{\mathbf{e}}_R\| \;[\si{\radian}]$ & $0.0032$ \\

        $\overline{\mathrm{MEI}} \; [-]$ & $0.7723$ &
        $\mathrm{RMS}_{\mathbf{e}_R} \; [\si{\radian}]$ & $0.0047$ \\
        \bottomrule
    \end{tabular}
\end{table}

\begin{figure}[t]
\centering
\subfloat[Free Flight\label{fig:subA}]{
    \includegraphics[width=0.45\linewidth]{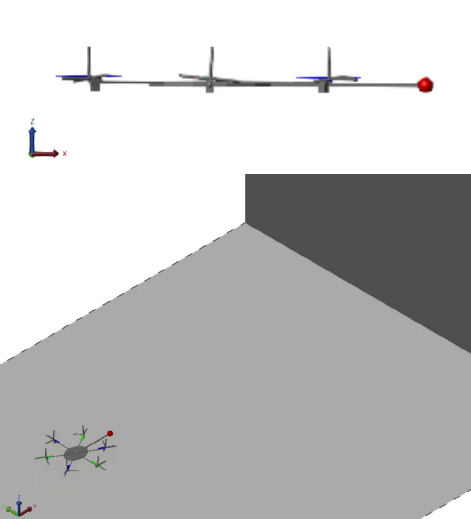}
}
\subfloat[Wall Interaction\label{fig:subB}]{
    \includegraphics[width=0.45\linewidth]{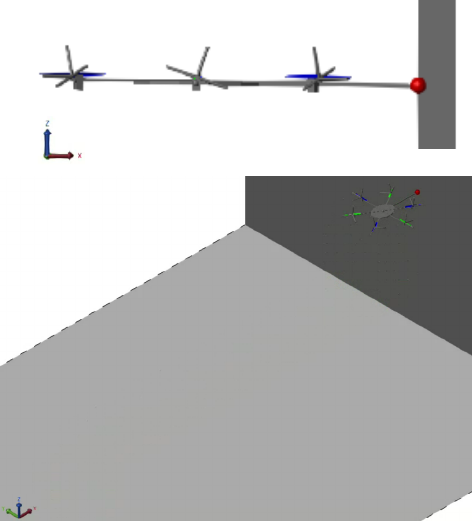}
}
\caption{Snapshots of the Simscape simulation (video in the supplementary material), showing the variation of the cant angle as the platform interacts with the wall.}
\label{fig:simscape_comp}
\end{figure}

Representative snapshots of the task execution are shown in Figure~\ref{fig:simscape_comp}, while the corresponding performance metrics are reported in Table~\ref{tab:simscape_kpis}. 
The results show that the proposed cant-angle selector enables the platform to complete the inspection sequence while adapting the tilt configuration to track the reference trajectory and counteract interaction forces.

Compared to the numerical simulations in Section~\ref{subsec:MC_analysis}, a moderate degradation in tracking accuracy is observed, especially during the transitions between free flight and contact, mainly due to the increased complexity of the physics-based model, which explicitly accounts for contact effects and more realistic actuation dynamics.
A further remark concerns the actuation-related KPIs. Both $\overline{\mathrm{FEI}}$ and $\overline{\mathrm{MEI}}$ are higher than in the numerical MC simulations due to the nature of the considered task. In particular, repetitive wall interaction induces a persistent force demand along the wall normal, thereby promoting more constructive thrust combinations and a more sustained actuation effort.
Nevertheless, the controller preserves platform stability and successfully completes the task.

Overall, these results confirm the feasibility of the proposed polytope-based cant-angle selector for complete interaction tasks, while highlighting possible improvements in transient response and interaction control.


\section{Conclusion}\label{sec:conclusion}
This paper presented a control framework for $\alpha$--TingH platforms performing in-contact tasks, featuring a lightweight polytope-based cant--angle selection strategy.  
The approach exploits an offline-computed LUT of zero-moment force polytopes to identify feasible cant angle candidates online, and then selects the optimal angle via a minimization that balances a low-tilt preference with temporal smoothness.  
Numerical validation results show that the proposed strategy improves pose-tracking accuracy and force efficiency compared to a baseline optimization-based allocation method, while substantially reducing the computational burden.
Furthermore, physics-based simulations carried out in Simscape demonstrate that the proposed framework remains effective in realistic interaction scenarios, enabling stable wall-contact tasks while maintaining satisfactory performance.

These results highlight the potential of geometric polytope-based reasoning for simplifying control allocation in thrust-vectoring multirotor platforms while preserving interaction performance. Future work will extend the method to account explicitly for interaction-induced moments and will include experimental validation on the real platform, as well as real-time implementation on embedded onboard hardware.


\bibliographystyle{IEEEtran}
\bibliography{bibliography}


\end{document}